\documentclass{article}

    \PassOptionsToPackage{numbers, compress}{natbib}
 \usepackage[preprint]{neurips_2026}

\usepackage[utf8]{inputenc} 
\usepackage[T1]{fontenc}    
\usepackage{hyperref}       
\usepackage{url}            
\usepackage{booktabs}       
\usepackage{amsfonts}       
\usepackage{nicefrac}       
\usepackage{microtype}      
\usepackage{subcaption}
\usepackage{multirow} 
\usepackage{tabularx} 
\usepackage{array}    
\usepackage[table]{xcolor}  
\usepackage{enumitem}
\usepackage{amsmath}
\usepackage{amssymb}
\usepackage{mathtools}
\usepackage{amsthm}
\usepackage[capitalize,noabbrev]{cleveref}

\usepackage{graphicx}
\usepackage{wrapfig}
\graphicspath{ {./figures/} }

\theoremstyle{plain}
\newtheorem{theorem}{Theorem}[section]

\newtheorem{lemma}[theorem]{Lemma}

\theoremstyle{definition}
\newtheorem{definition}[theorem]{Definition}
\newtheorem{assumption}[theorem]{Assumption}
\theoremstyle{remark}
\newtheorem{remark}[theorem]{Remark}

\definecolor{qLow}{RGB}{232,245,233}   
\definecolor{qMid}{RGB}{255,248,225}   
\definecolor{qHigh}{RGB}{255,235,238}  
\definecolor{qTail}{RGB}{245,245,245}  

\definecolor{rank1}{HTML}{BDD7EE}
\definecolor{rank2}{HTML}{DDEBF7}
\definecolor{rank3}{HTML}{F2F2F2}
\definecolor{rank4}{HTML}{FFFFFF}

\newcolumntype{P}[1]{>{\raggedright\arraybackslash}p{#1}}
\newcolumntype{Y}{>{\raggedright\arraybackslash\ttfamily}X}

\title{Curriculum Learning for LLM Pretraining \\
  An Analysis of Learning Dynamics}

\author{%
  Mohamed Elgaar$^1$ \quad
  Hadi Amiri$^1$ \\ 
  $^1$University of Massachusetts Lowell \\
  \texttt{\{melgaar, hadi\}@cs.uml.edu} \\
}

\begin{document}

\maketitle

\begin{abstract}
Curriculum learning changes the order of pretraining data, but it remains unclear how ordering changes the learning dynamics. We pretrain models from 14M to 1B parameters for 300B tokens under three linguistically motivated curricula---Age-of-Acquisition, word frequency, and Verb Variation (VV)---and compare each against Random ordering. We analyze latent training phases, gradient noise scale (GNS), and the singular-value structure of the output head. We find that training follows a shared sequence of latent phases, while curricula mainly change time spent in each phase. Random ordering yields higher GNS at 14M--70M and late singular-entropy spikes up to 160M, consistent with noisier gradients and output-head saturation. A reverse-order VV control shows that direction matters: descending order loses much of the accuracy advantage of the ascending curriculum. At larger scales, these stability differences are smaller. These results indicate that the curricula studied here are associated with more stable within-phase training in smaller models rather than with the creation of new phases.
\end{abstract}

\section{Introduction}

Pretraining large language models requires substantial compute~\citep{kaplan2020scaling, hoffmann2022training}. Accordingly, modern practice typically trains for a single epoch over massive corpora: revisiting data is inefficient when new data are abundant~\citep{komatsuzaki2019one}. Compute-optimal scaling laws similarly treat the budget as the total number of tokens consumed from a unique corpus~\citep{hoffmann2022training}. In this single-pass setting, data order can affect which examples shape each training stage, because each sample is encountered exactly once. Curriculum learning---presenting training examples in a structured order, often easy-to-hard---changes which samples appear early or late in this stream~\citep{bengio2009curriculum, elman1993learning}. Such ordering encodes a natural hypothesis: learning may be easier when foundational patterns precede rarer or more complex ones.

Despite this intuition, results on curriculum learning for LLM pretraining are mixed. Some studies report gains at fixed compute budgets~\citep{fan2023irreducible}, whereas others find diminishing or negligible effects at scale~\citep{campos2021curriculum}. Mixed evidence leaves a mechanistic ambiguity: curricula could change the sequence of phases a model traverses during training, or primarily change which data the model sees at each phase. To design curricula that improve training consistently, we need to separate these possibilities.

We focus on smaller models because capacity constraints make ordering effects easier to detect. These models face structural limits such as the softmax bottleneck~\citep{godeysmall, michaelov-bergen-2023-emergent}: the output head can have too little effective rank to represent the high-rank distributions required for language modeling. As training progresses, this mismatch can push the output head toward a saturated spectral regime, where its singular-value structure becomes increasingly dominated by the capacity limits of the head rather than by useful representational growth. We refer to this late-training degeneration as output-head spectral saturation. It matters for curriculum learning because data ordering changes which examples a capacity-constrained model encounters as it approaches this limit; a curriculum may help by keeping late-phase optimization stable, even if it does not change the sequence of broad learning phases.

\begin{figure*}[t]
    \centering
    \includegraphics[width=0.9\linewidth]{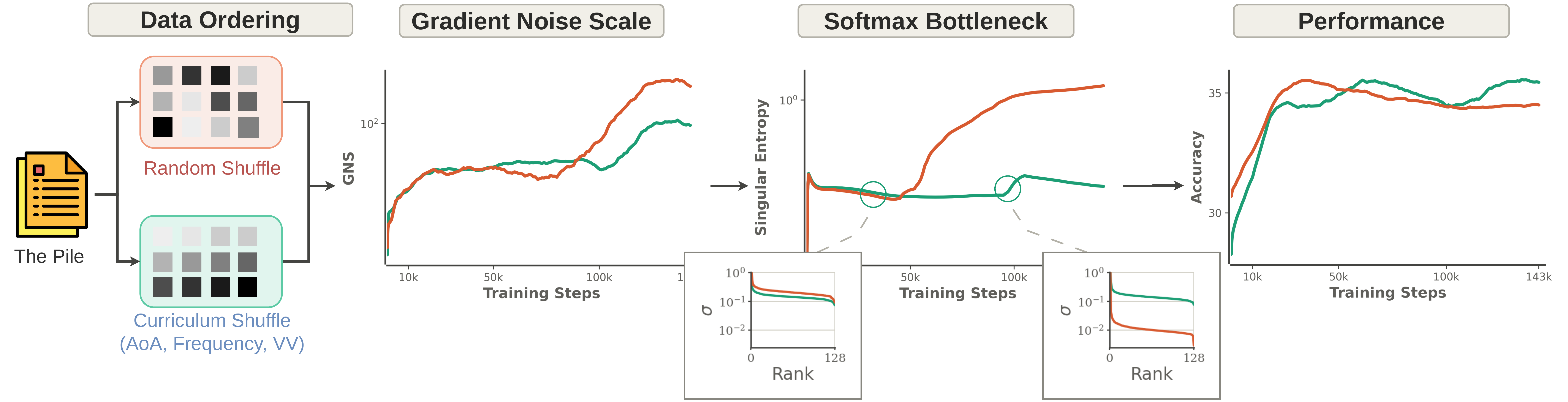}
    \caption{Overview of the data-ordering comparison and diagnostics. Starting from the same fixed Pile samples, we compare Random ordering against curricula based on Age-of-Acquisition, word frequency, and Verb Variation. The remaining panels summarize the diagnostics used throughout the paper: gradient noise scale, output-head singular entropy and spectrum shape, and evaluation accuracy. In smaller models, Random ordering shows higher late-training GNS and stronger spectral concentration, while the linguistic curricula keep these stability diagnostics lower and can preserve accuracy on selected evaluations later in training.}
    \label{fig:overview}
\end{figure*}

We test these questions by training Pythia models (14M--410M parameters) for 300B tokens under three linguistically motivated curricula---Age-of-Acquisition, word frequency, and Verb Variation (VV)---and comparing them against Random ordering; for 1B parameters, we compare Random and VV due to compute constraints. Beyond aggregate metrics, we analyze learning dynamics in two ways. First, we perform latent phase analysis by fitting Hidden Markov Models to training trajectories. Second, we track two stability diagnostics. The gradient noise scale~\citep{mccandlish2018empirical} measures how noisy stochastic gradients are relative to their signal. Singular entropy summarizes the singular-value structure of the language-modeling head and serves as a diagnostic for output-head saturation~\citep{godeysmall}. In our late-training regime, the concerning pattern is a sharp increase in this diagnostic under Random ordering; lower values at the same training step indicate that the output head has avoided this saturation-associated spike, not that lower singular entropy is universally desirable. Figure~\ref{fig:overview} summarizes the comparison: the same Pile samples are randomly shuffled or sorted by linguistic scores, then analyzed through GNS, output-head spectral saturation, and accuracy.

For smaller models, training follows a shared sequence of latent phases. Curricula mainly change which data appears within each phase. The effect is strongest at small scales: Random ordering exhibits higher GNS at 14M--70M and stronger late-training spectral saturation up to 160M. The linguistic curricula reduce these diagnostics at matched compute. By 410M, the differences narrow.

This paper makes the following contributions:
\begin{itemize}[leftmargin=*]
    \item We show that pretraining trajectories under different data orderings share a consistent latent phase structure. Using an HMM analysis across orderings, we find that curricula mainly change phase occupancy and within-phase data exposure rather than changing the phase sequence.
    \item We find that linguistic curricula reduce gradient noise scale and output-head spectral saturation in smaller models. A reverse-order VV control loses much of the corresponding accuracy gain, showing that ordering direction matters. We interpret the GNS reductions using a standard variance-control view of SGD, and we analyze singular entropy as a separate diagnostic of softmax-bottleneck saturation.
    \item We report experiments spanning five model sizes (14M--410M) trained for 300B tokens under three linguistic curricula and Random ordering, with additional 1B runs for Random and VV, showing that the measured stability effects are scale-dependent and most pronounced for capacity-constrained models.
    \item We separate ideal difficulty from operational linguistic scores: AoA, frequency, and VV correlate with empirical sample loss, but this loss is only a proxy for the ideal score and the same orderings also induce domain schedules in The Pile. The observed effects should therefore be interpreted as ordering effects rather than isolated interventions on linguistic difficulty alone.
\end{itemize}

\section{Related Work}
\label{sec:related_work}

This work relates to curriculum learning for language model pretraining, analyses of training dynamics, and scaling effects. Curriculum learning orders examples to shape optimization, classically in easy-to-hard form~\citep{bengio2009curriculum, elman1993learning}. While many curriculum formulations and schedules have been proposed~\citep{cornacchia2023mathematical, kong2021adaptive, zhou2018minimax}, their implications are less clear in single-pass LLM pretraining. Empirical findings are mixed: linguistically motivated orderings can show diminishing returns at scale~\citep{campos2021curriculum}, whereas proxy-learnability orderings can improve perplexity and downstream accuracy for larger models~\citep{fan2023irreducible}; length-based curricula may affect stability with uncertain effects on final quality~\citep{li2022stability, nagatsuka2023length}. A closely related study evaluates vanilla, pacing-based, and interleaved curricula for LLM pretraining, with main 0.5B and 1B runs trained for 10B and 20B tokens, respectively~\citep{zhang2026randomsamplingefficientlanguage}. Our 300B-token runs study a later training regime, where ordering effects can appear as stability differences rather than only early- and mid-training efficiency gains.

Llama 3 reports in-run changes to its pretraining mix~\citep{llama3herd}. OLMo 2 splits base training into a long pretraining stage and a shorter mid-training stage, which upsamples high-quality sources~\citep{olmo2furious}. Olmo 3 extends this staged recipe with pretraining, midtraining, and long-context stages~\citep{olmo3}. These systems motivate treating mixture exposure as part of the training procedure. Our experiments isolate a narrower intervention: the sample set is fixed, and sorting by linguistic scores changes when domains appear without changing which samples are included.

Data Mixing Laws predicts the effect of fixed mixture proportions, framing dynamic schedules as a promising extension rather than implementing online mixture adjustment~\citep{dataMixingLaws}. Olmix studies the offline mixing schema as domain sets evolve during language-model development, while leaving online methods that adjust mixture weights during the training run to future work~\citep{olmix}. These studies separate mixture choice from ordering; our setup changes ordering while holding samples fixed, inducing a source-domain schedule as a byproduct rather than optimizing mixture weights directly.

To capture within-training changes that aggregate metrics obscure~\citep{kaplan2020scaling, hoffmann2022training}, we model training trajectories with Hidden Markov Models. Prior approaches analyze trajectories via data attribution~\citep{koh2017understanding, pruthi2020estimating} or simulate sequence effects on metric evolution under different data orderings~\citep{guu2023simfluence, chai2024training}. HMMs have been used to identify consistent learning phases across random seeds~\citep{wal2025polypythias}; we extend this by fitting a joint HMM across orderings to test whether curricula change phase structure or primarily reallocate exposure within shared phases.

To study possible mechanisms, we focus on two diagnostics with different roles. Gradient noise scale provides an optimization-stability lens~\citep{mccandlish2018empirical}; we use a stylized variance-control argument to motivate why score-based pacing could reduce this quantity when the score tracks gradient variance. Singular entropy, by contrast, is a capacity diagnostic grounded in the softmax bottleneck: small models can saturate or degrade late in training due to limited-rank output heads~\citep{godeysmall, michaelov-bergen-2023-emergent}. Our experiments test whether ordering changes phase structure or instead changes within-phase exposure and the stability diagnostics associated with those phases.

\section{A Variance-Control Lens for Curriculum Learning} 
\label{sec:theoretical_framework}
This section uses a simplified SGD variance argument to interpret one measurable implication of curriculum learning: changes in stochastic-gradient variance. The goal is not to derive an end-to-end theory of deep-network pretraining, non-convex optimization, Adam, or output-head saturation. Instead, we use a standard SGD variance bound as an interpretive device for the empirical prediction that sorted curricula can reduce gradient noise scale when they delay exposure to higher-variance examples. Appendix~\ref{sec:appendix_theory_proofs} gives the formal setup, assumptions, pacing definition, and proofs.

\paragraph{The Softmax Bottleneck and Rank Saturation}
Small models can exhibit late-stage saturation effects due to the softmax bottleneck~\citep{godeysmall}. Their output head can have too little effective rank to represent the high-rank distributions required for language modeling, creating a capacity-constrained regime in which the spectrum of the language-modeling head becomes increasingly concentrated. We track this output-head spectral degeneration using singular entropy in \S\ref{sec:results} (Figure~\ref{fig:entropy}). This diagnostic is separate from the convex variance-control sketch: the sketch motivates GNS, while singular entropy measures whether the language-modeling head enters a concentrated spectral regime.

\paragraph{The Link Between Ideal Difficulty and Gradient Variance}

\citet{hacohen2019power} advocate training first on examples with low loss under the optimal model $\theta^*$. Following this static notion of difficulty, we assume that examples with higher ideal difficulty can also have higher stochastic-gradient variance near $\theta^*$. In our experiments, $\theta^*$ is unobserved, so we use average per-sample loss under trained checkpoints as an empirical proxy for ideal difficulty rather than as a direct measurement of it.

\paragraph{The Gradient Noise Scale}

The gradient noise scale (GNS) is a measure of the signal-to-noise ratio in stochastic gradient estimates~\citep{mccandlish2018empirical}. Let $G_t = \mathbb{E}[g_t \mid \theta_t] = \nabla F(\theta_t)$ denote the true gradient at step $t$ and $\Sigma_t = \mathrm{cov}(g_t \mid \theta_t)$ the covariance of the stochastic gradient. The simplified noise scale is defined as
\begin{equation}
    \mathcal{B}_t \;=\; \frac{\mathrm{tr}(\Sigma_t)}{\|G_t\|_2^2},
    \label{eq:gns}
\end{equation}
which measures the total gradient variance relative to the squared gradient magnitude. When $\mathcal{B}_t$ is large, stochastic gradient estimates are noisier relative to their signal; when $\mathcal{B}_t$ is small, gradients are more consistent across batches.

For a sampling distribution $\mathcal{P}_t$ with effective gradient variance $\sigma_t^2$, the noise scale satisfies $\mathcal{B}_t \propto \sigma_t^2 / \|G_t\|_2^2$. If sorting changes the effective variance of local training windows, it can change $\mathcal{B}_t$. This motivates a measurable prediction: curriculum-based orderings should have lower measured GNS than Random ordering when early sorted windows are lower-variance than the full stream and gradient magnitudes are otherwise comparable. We examine this prediction indirectly in \S\ref{sec:results} using GNS trajectories, while noting that we do not directly measure the per-window gradient variance assumed by the stylized argument.

\paragraph{A Standard Stability Bound as Motivation}

Standard SGD analysis shows that under strong convexity and Lipschitz gradients, the asymptotic stability radius scales with gradient variance $\sigma^2$: sampling schemes with bounded variance yield tighter stability bounds~\citep{bottou2018optimization,garrigos2023handbook}. This motivates controlling variance via score-based pacing when the score tracks variance-relevant difficulty.

\begin{remark}[Curriculum learning and variance control] \label{rem:variance_control}
In an idealized strongly convex setting, the stability radius of SGD scales with the effective gradient variance: sampling schemes that keep $\sigma_t^2$ bounded yield tighter stability bounds. Under uniform sampling, $\sigma_t^2$ may drift toward a high-variance regime late in training; sorted curricula induced by a variance-aligned score delay this drift by placing lower-score windows early in the one-pass stream and higher-score windows later.
\end{remark}

\paragraph{Mapping the framework to our curricula}
\label{sec:theory_to_practice}
Each curriculum defines a deterministic ordering over the one-epoch pretraining stream by sorting fixed-length 2048-token samples (Section~\ref{sec:method}). We map the framework to this setting by treating the curriculum score as $d(z)$ and the fully sorted stream as linear rank pacing: early batches come from lower-score windows and later batches from higher-score windows. Random ordering instead approximates a time-invariant distribution over the same fixed samples. The empirical test is therefore weaker than the theory: the linguistic curricula are expected to have lower measured GNS than Random ordering if their score windows reduce effective gradient variance, but the experiments do not identify variance, linguistic difficulty, and domain mixture separately.

\begin{figure*}[t]
    \centering
    \includegraphics[width=\linewidth]{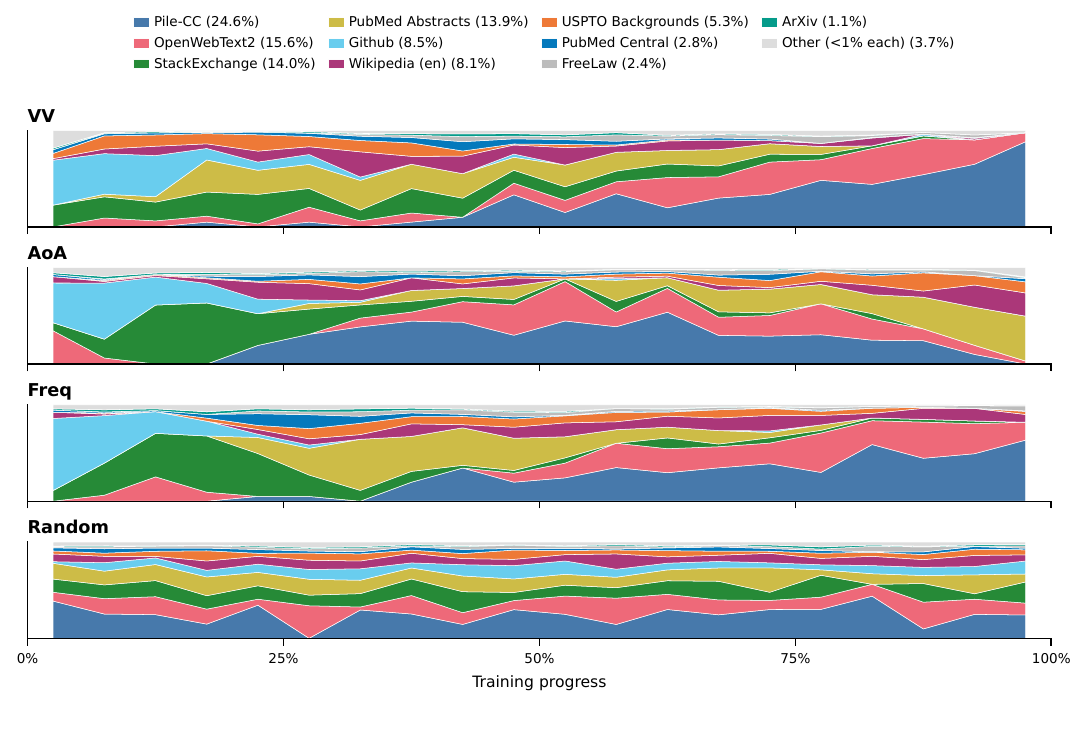}
    \caption{Source-domain composition over training for each data ordering. Each panel shows the fraction of training samples from each Pile component across normalized training progress; legend values give the overall corpus share. Random preserves a roughly stationary mixture, whereas the linguistic curricula induce different domain schedules even though all orderings contain the same fixed 2048-token samples.}
    \label{fig:data_mix}
\end{figure*}

\section{Pretraining Curriculum Design}
\label{sec:method}
The curriculum operates at the level of individual training samples. Following the Pythia data processing pipeline~\citep{biderman2023pythia}, documents from The Pile~\citep{gao2020pile} are partitioned into fixed-size, 2048-token samples. We study a controlled setting in which this set of 2048-token samples is held fixed and only the ordering of samples is changed; observed differences therefore reflect the ordering intervention rather than changes in the underlying sample set.

Each curriculum assigns a scalar score to every sample and sorts samples accordingly. In line with \Cref{def:diff_score}, we use average per-sample loss under trained checkpoints as an empirical proxy for ideal difficulty, since the optimal model $\theta^*$ is unobserved. We study three inexpensive linguistic indices: Age-of-Acquisition (AoA)~\citep{kuperman2012age}, word frequency using SUBTLEX$_{\text{US}}$ and the Zipf scale~\citep{brysbaert2009subtlex, van2014subtlex}, and Verb Variation (VV)~\citep{guiraud1960problèmes}. Across model scales, these indices correlate positively with sample loss (average Pearson correlations: VV 0.764, frequency 0.730, AoA 0.616), so the scores are aligned with this empirical proxy in our setting. These indices capture complementary linguistic signals: AoA reflects acquisition norms, frequency reflects lexical distributional statistics, and VV reflects verb-type diversity. We sort samples in ascending score. As a directionality control, we also compare VV against a descending-score order for a representative small-model run, exposing high-score samples before low-score samples while keeping the same fixed sample set and training budget. Details of scoring functions, correlation analyses, and qualitative excerpts appear in Appendix~\ref{sec:curriculum_details}.

\subsection{Linguistic Characterization of Curriculum Indices}
\label{sec:linguistic_characterization}

Qualitative inspection of curriculum quantiles shows recurring content differences. Low-score regions (low frequency, low AoA, low VV) tend to contain code, structured data, and specialized vocabulary; mid-range samples include expository and technical prose; high-score tails can contain repetitive or degenerate content (e.g., keyword lists, verb lexicons). VV should be interpreted as detected verb-type diversity rather than a pure measure of semantic complexity, since verb-list formats can dominate the extreme tail. These patterns provide context for probe-level differences in \S\ref{sec:results}.

Because The Pile is heterogeneous, sorting by these linguistic scores also changes the source-domain mixture encountered over training. Figure~\ref{fig:data_mix} plots the source composition for each ordering as a function of normalized training progress. Random ordering remains close to the global mixture throughout, while VV, AoA, and Frequency create distinct domain schedules: for example, Frequency places more GitHub and StackExchange content early, more PubMed Abstracts near the middle, and more Pile-CC and OpenWebText2 content late. Domain-level analyses show that this entanglement is not incidental: Pile domains differ in empirical sample loss and in their average linguistic scores. We therefore interpret the curricula as linguistic orderings over a heterogeneous corpus, not as isolated interventions on a single linguistic property.

\subsection{Models and Training Setup}
\label{sec:training_setup}
The experiments use the Pythia suite~\citep{biderman2023pythia}, specifically models with 14M, 31M, 70M, 160M, 410M, and 1B parameters. Pythia provides a standardized architecture and logging across model sizes, so data-ordering comparisons use the same architecture and logs; we also compare against stability analyses such as PolyPythias~\citep{wal2025polypythias}. The training setup replicates the original Pythia configuration, including learning rate, batch size, and optimizer, and these hyperparameters are held fixed across orderings for comparability.

For model sizes 14M--410M, we train separate runs from scratch under four primary orderings: Random ordering (a single random shuffle of The Pile) and three ascending lexical-score curricula induced by Age-of-Acquisition, word frequency, and VV. Models are trained for 20B, 60B, and 300B tokens, and we analyze intermediate checkpoints at 10B tokens where available. These durations allow observation of dynamics within and across key learning phases identified in prior stability research (e.g., an initial phase typically concluding around 20B tokens, and a development phase extending from 20B to around 200B tokens~\citep{wal2025polypythias}). For 1B parameters, due to compute constraints, we train and evaluate only Random and VV.

Detailed computational costs are provided in Appendix~\ref{sec:compute_costs}. Training the 70M, 160M, and 410M models consumed approximately 530, 1,140, and 2,730 A100 GPU-hours, respectively. The 300B-token runs are chosen to observe late-stage behavior, enable the HMM to observe complete phase sequences, and assess final performance in a realistic single-epoch regime; shorter runs would suppress the saturation effects central to our hypothesis.

\subsection{Analyzing Learning Dynamics}
\label{sec:hmm_methodology}
The influence of curricula on learning is examined from two perspectives: the evolution of external linguistic capabilities and the internal state of the model.

\paragraph{HMM-based Identification of Learning Phases}
Capability visualizations indicate what the model learns but provide limited insight into training stability and internal state transitions. We model training trajectories with a Hidden Markov Model (HMM) following~\citet{hu2023latent} to obtain a low-dimensional, discrete training map that facilitates identification of common learning phases and analysis of how curricula affect traversal of these phases.

Direct HMM fitting on the high-dimensional weight space is computationally infeasible. Instead, we compute 14 checkpoint-level metrics from the model's weights, standardize the resulting time series, and use them as observation sequences; Appendix~\ref{sec:hmm_metrics} lists these metrics and the state-count sweep. The metrics capture properties of the weight distribution (e.g., $L_1$ and $L_2$ norms) and the function computed by each layer (e.g., singular values).

To ensure the HMM-based phase analysis reflects stable learning dynamics rather than idiosyncrasies of a single run, we train three seeds for each approach under the 14M and 31M models. These extra replicates let us test whether curriculum orderings versus Random ordering exhibit distinct but internally consistent latent-state sequence patterns across seeds.

We fit a Gaussian HMM to these sequences using the standard expectation-maximization procedure (Baum-Welch). As in \citet{wal2025polypythias}, we use five latent states, which supports direct comparisons across orderings. We train a single HMM jointly across the Age-of-Acquisition, Frequency, Verb Variation, and Random orderings so that phases are directly comparable.

The trained HMM provides a state transition diagram that functions as a training map and reveals a sequence of latent learning phases shared across orderings. The subsequent analysis assesses whether curricula change which states are visited or instead alter the data processed within shared states, which we compare against GNS, singular entropy, and accuracy.

\paragraph{Evaluation Benchmarks}
The analysis relies on visualizations that track multiple capability-specific metrics throughout training. To capture the development of diverse linguistic and reasoning abilities, each model checkpoint is evaluated on benchmarks covering question answering, long-range context comprehension, logical reasoning, and commonsense understanding. Descriptions appear in Appendix~\ref{sec:probed_metrics_details}.

\begin{figure*}[t]
    \centering
    \includegraphics[width=\linewidth]{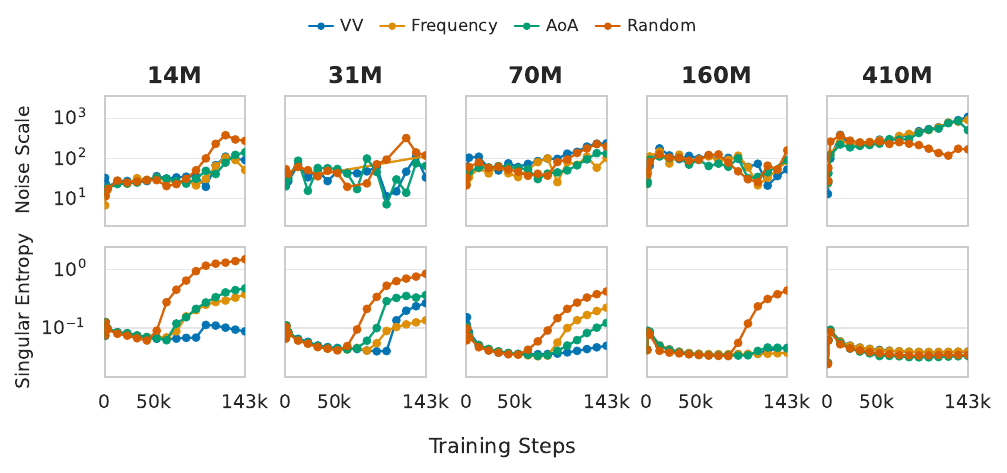}
    \caption{Training-stability diagnostics across model sizes and data orderings. \textbf{Top:} gradient noise scale (GNS) versus training step. Random ordering tends to produce higher GNS for smaller models (14M--70M), indicating noisier gradient estimates relative to their signal; at larger scales (160M--410M), the gap is smaller and less consistent. \textbf{Bottom:} singular entropy of the language modeling head versus training step. For models up to 160M parameters, Random ordering produces sharp late-stage increases in singular entropy, indicating movement toward a spiked spectrum associated with saturation; ascending lexical-score curricula (AoA, frequency, VV) maintain lower entropy and do not show the same late spike.}
    \label{fig:gns}
    \label{fig:entropy}
\end{figure*}

\section{Results} \label{sec:results}

\subsection{Ordering effects can be heterogeneous across capabilities}
Using BLiMP probes at 300B tokens, we observe ordering-dependent differences that are concentrated in a small number of syntactic phenomena rather than spread uniformly across probes.
Across 14M--410M models, the Verb Variation curriculum (VV) improves \emph{wh}-questions object-gap accuracy by \(\approx\)9 percentage points on average at the end of training, with the largest gains at 14M--70M and no benefit at 410M; it also improves causatives by \(\approx\)4--5 points on average.
In contrast, the Frequency curriculum underperforms Random on superlative quantifiers by \(\approx\)14 points on average, with additional average drops on ellipsis-N-bar (\(\approx\)3 points) and only-NPI scope (\(\approx\)4 points), though these effects vary by scale.
Age-of-Acquisition shows no consistent cross-probe BLiMP advantages or regressions relative to Random under our thresholds, suggesting these effects are ordering-specific rather than a general consequence of non-random data presentation.

Ordering effects are sparse and sometimes negative: VV helps selected predicate-argument and filler-gap probes, Frequency hurts selected quantifier and NPI probes. The main result is therefore not that linguistic curricula improve downstream accuracy globally, but that data order can selectively preserve or degrade specific capabilities while changing training stability.

The probe-level differences are consistent with the text characteristics in \S\ref{sec:linguistic_characterization}. VV explicitly prioritizes samples with higher verb diversity, increasing early exposure to varied predicate--argument structures and clause-level dependencies; one possible explanation is that this supports generalization on filler-gap constructions (\emph{wh}-movement) and argument-structure alternations (causatives). Frequency, by contrast, places low-frequency, highly structured content (e.g., code and templated text) early in training; this can reduce early exposure to the natural-language contexts that most strongly evidence quantifier scope and NPI licensing patterns. Appendix~\ref{sec:appendix_shuffle_blimp} provides the full data and visualizes these trajectories. The stability diagnostics below are measured along the checkpoint trajectories.

\begin{figure*}[t]
    \centering
    \includegraphics[width=\linewidth]{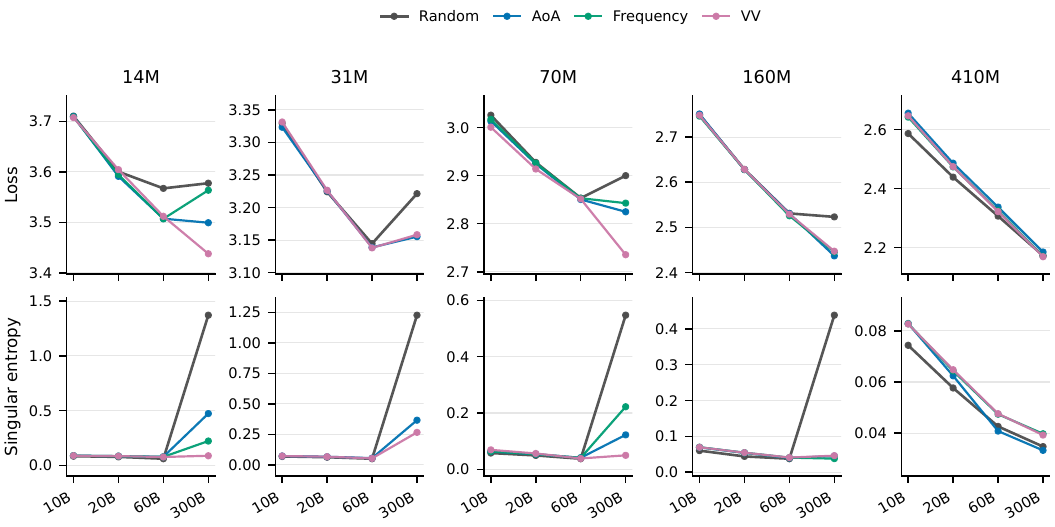}
    \caption{Effect of training token budget on loss and singular entropy across model sizes. The late-stage singular-entropy behavior differs by scale and ordering. At 300B tokens, Random exhibits large singular-entropy increases for smaller models, while the linguistic curricula remain lower; this gap largely disappears at 410M. The pink curve corresponds to Verb Variation (VV).}
    \label{fig:token_budget_effect}
\end{figure*}

\subsection{Linguistic curricula reduce gradient noise scale and singular entropy in smaller models}
The linguistic curricula reduce gradient noise scale for smaller models, consistent with the observable pattern motivated by the variance-control lens in \S\ref{sec:theoretical_framework}. GNS is estimated using 32 batches of size 32. Figure~\ref{fig:gns} shows that for models at 14M--70M scale, Random ordering often exhibits higher GNS than the linguistic curricula, particularly in later training phases. Higher GNS under Random ordering indicates noisier gradient estimates~\citep{mccandlish2018empirical}, which can make late-phase optimization less stable. Because the experiments measure GNS rather than per-window gradient variance, this result should be interpreted as support for the narrower stability pattern, not as a direct validation of the idealized difficulty-variance assumption. At larger scales (160M--410M), the GNS differences between orderings are smaller and less consistent.

Random ordering also produces sharp late-stage increases in singular entropy for models up to 160M parameters (Figure~\ref{fig:entropy}), indicating movement toward a spiked singular value distribution associated with saturation~\citep{godeysmall}. The linguistic curricula maintain lower singular entropy throughout training and do not show the same late spike. The 410M model shows minimal entropy differences across orderings, consistent with the broader capacity interpretation: spectral saturation matters most in smaller models, while larger models are less sensitive to these lexical orderings. The additional 1B Random--VV comparison follows this scale trend: the measured differences are smaller than the 14M--160M effects, so we do not treat 1B as evidence for a persistent large-scale curriculum advantage.

The token-budget sweep in Figure~\ref{fig:token_budget_effect} shows why the 300B-token horizon changes the comparison. At 10B--60B tokens, loss improves for all orderings and singular entropy remains low for most small-model runs, so the orderings appear similar under the shorter budgets used in prior curriculum-pretraining studies~\citep{zhang2026randomsamplingefficientlanguage}. However, as training continues toward 300B tokens, the loss for the Random ordering improves initially but then worsens later in training. Coinciding with this loss degradation, Random develops large singular-entropy spikes for 14M--160M models, while the linguistic curricula stay substantially lower. Extended training therefore reveals late-stage spectral saturation and scale-dependent stability effects that shorter-horizon studies were not designed to observe.

The reverse-order control in Appendix~\ref{sec:appendix_reverse_control} tests whether ordering direction matters. Reversing VV reduces accuracy relative to the ascending curriculum throughout most of training and removes the late-training accuracy advantage of the ascending score order. The entropy result is mixed: descending order rises earlier and higher than ascending order, but remains well below Random at the end of training. We therefore treat the reverse run as evidence that ordering direction affects the stability-performance tradeoff for this score, not as a complete validation of the variance-control lens or a clean separation from domain-mixture effects.

\subsection{Shared phases; within-phase exposure drives stability}

\begin{wrapfigure}{r}{0.48\textwidth}
    \centering
        \begin{subfigure}{\linewidth}
            \centering
            \includegraphics[width=\linewidth]{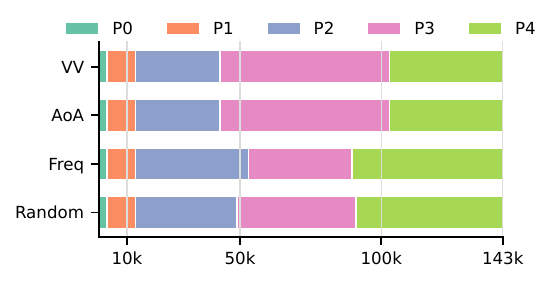}
            \caption{Mean training steps spent in each HMM phase by ordering.}
            \label{fig:hmm_time}
        \end{subfigure}\\[0.75ex]
        \begin{subfigure}{\linewidth}
            \centering
            \includegraphics[width=\linewidth]{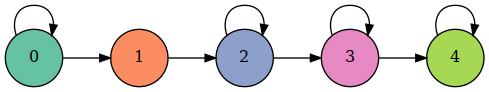}
            \caption{HMM state transition diagram.}
            \label{fig:hmm_transitions}
        \end{subfigure}
    \caption{Shared phases and within-phase dynamics. \subref{fig:hmm_time} shows the mean number of training steps assigned to each HMM phase for Random, Age-of-Acquisition, Frequency, and VV. \subref{fig:hmm_transitions} shows the shared HMM state transition diagram learned jointly across orderings..}
    \vspace{-20pt}
    \label{fig:hmm_overview}
\end{wrapfigure}

We train a single HMM jointly over 14M trajectories from Age-of-Acquisition, Frequency, Verb Variation, and Random, yielding a shared set of latent training phases (Figure~\ref{fig:hmm_overview}). This mirrors the finding of \citet{wal2025polypythias} that Pythia pretraining proceeds through consistent phases across random seeds when training the same architecture on the same underlying corpus. In our setting, we hold the architecture, optimizer configuration, and data source fixed and change only the ordering; the shared transition map across orderings indicates that curricula do not create new macro-phases under this five-state model. Instead, the orderings differ in phase occupancy: Frequency spends more time in phase P2, while AoA and VV spend more time in P3 before entering the final phase.

Polypythias~\citep{wal2025polypythias} interpret these phases in terms of broader training phenomena: an early learning phase (roughly $10^3$--$10^4$ steps) where representations change rapidly and linguistic information begins to be encoded, followed by a critical learning phase (roughly $10^4$--$10^5$ steps) in which most benchmark improvements occur and representations stabilize, and then a later regime in which progress slows and small models can exhibit saturation and instability. Under this view, our result that orderings share a transition structure but differ in time spent in each state suggests a mechanism for the observed ordering effects: the curricula change which data is encountered during the early and critical phases and how long the model remains there, which in turn can affect late-phase behavior.

Across orderings, the HMM states are shared; the differences appear in phase occupancy and phase-conditional exposure. The resulting GNS and singular-entropy differences then appear as late-phase stability effects in smaller models. This phase result constrains the interpretation of the stability gains. If curricula were inducing qualitatively different learning trajectories, we would expect ordering-specific transition structures rather than a shared HMM map. Instead, the evidence points to a weaker but more actionable effect: the same coarse training program is traversed under different data exposures and residence times. The GNS and singular-entropy gaps should therefore be interpreted as within-phase stability differences, not as evidence that the curricula create new developmental stages.

\section{Conclusion} \label{sec:conclusion}
We studied how pretraining data order affects learning dynamics by training Pythia models under Random ordering and three linguistic curricula.

Across orderings, the HMM analysis finds a shared sequence of latent phases; curricula change phase occupancy and phase-conditional exposure rather than creating new macro-phases. In smaller models, Random ordering produces higher GNS and stronger late-stage singular-entropy spikes, while the linguistic curricula keep these stability diagnostics lower. The gaps narrow at larger scales, suggesting that these ordering effects matter most when model capacity is constrained and late-stage saturation is a concern.

The main limitations are that all runs use a fixed set of 2048-token sequences, the linguistic scores are imperfect proxies for ideal difficulty, and sorting by these scores also induces domain schedules in The Pile. As a result, these experiments identify ordering effects over a heterogeneous corpus, but do not separate linguistic difficulty, source-domain exposure, and their interaction. In these Pythia runs, data ordering changes within-phase exposure and is associated with measurable stability differences.

\bibliographystyle{unsrtnat}
\bibliography{anthology_part1,anthology_part2,custom}


\newpage
\appendix

\section{Theoretical Framework: Additional Details and Proofs}
\label{sec:appendix_theory_proofs}

\subsection{Formal setup}
The training objective is modeled as a population risk
\[
    F(\theta) \;=\; \mathbb{E}_{z \sim \mathcal{P}}[\ell(\theta,z)],
\]
where $\theta \in \mathbb{R}^{d_\theta}$ are the model parameters, $z$ is a training example drawn from a data distribution $\mathcal{P}$, and $\ell$ is the per-example loss. SGD with a possibly time-varying sampling distribution $\mathcal{P}_t$ performs updates
\begin{equation}
    \theta_{t+1} \;=\; \theta_t - \eta \, g_t,
    \qquad
    g_t \;=\; \nabla_\theta \ell(\theta_t, z_t),
    \quad
    z_t \sim \mathcal{P}_t,
    \label{eq:sgd_update}
\end{equation}
with constant step size $\eta > 0$. The bound below uses standard convex-optimization assumptions, so it should be read as motivation for the GNS measurements rather than as a quantitative theory of LLM pretraining.

\subsection{Softmax bottleneck and singular entropy}
We state the formal lemma and define the diagnostic used in \S\ref{sec:theoretical_framework}.

The saturation problem follows~\citet{godeysmall}. Let $r$ denote the dimension of the final-layer representation (equivalently, an upper bound on the rank of the output head). The final linear layer maps an $r$-dimensional representation to a distribution over a vocabulary of size $V$, which can be viewed as learning a low-rank approximation $W_r \in \mathbb{R}^{V \times r}$ of a theoretically optimal, high-rank matrix $W^*$.

\begin{lemma}[Inherent Performance Gap, from~\citet{godeysmall}]
Let $W^*$ be the optimal language modeling head, with singular values $\{\sigma_i\}_{i=1}^V$. Any model with a linear head of rank at most $r$ incurs a rank-constrained approximation error equal to the discarded spectral energy:
\begin{equation}
    \min_{\text{rank}(W_r) \le r} \|W_r - W^*\|_F^2 = \sum_{i=r+1}^{V} \sigma_i^2
\end{equation}
When $r$ is small relative to the intrinsic rank of $W^*$, this approximation gap can be substantial and can produce saturation-related loss.
\end{lemma}

To quantify the spectral structure of the language modeling head during training, we use singular entropy as a diagnostic~\citep{godeysmall}. Let $\{\sigma_i\}_{i=1}^r$ denote the singular values of the head matrix $W$, normalized to form a probability distribution $p_i = \sigma_i / \sum_j \sigma_j$. The singular entropy is defined as the Kullback-Leibler divergence between this distribution and the uniform distribution:
\begin{equation}
    H_{\text{sing}}(W) \;=\; D_{\text{KL}}\left( p \,\|\, \mathcal{U} \right) \;=\; \sum_{i=1}^{r} p_i \log(r \cdot p_i),
    \label{eq:singular_entropy}
\end{equation}
where $\mathcal{U}$ is the uniform distribution over $r$ components. Note that $H_{\text{sing}}(W)=\log r - H(p)$, where $H(p)=-\sum_i p_i \log p_i$ is the Shannon entropy of $p$; thus $H_{\text{sing}}$ measures divergence from the maximum-entropy (uniform) spectrum and equals $0$ when $p$ is uniform. Low singular entropy corresponds to a more uniform distribution of singular values; high singular entropy corresponds to a more concentrated (spiked) spectrum. \citet{godeysmall} associate these spectral regimes with saturation dynamics.

\subsection{Ideal difficulty and gradient variance}
We state the formal assumption used in \S\ref{sec:theoretical_framework}.

\begin{definition}[Ideal Difficulty Score]\label{def:diff_score}
Following~\citet{weinshall2018curriculum}, the \emph{ideal difficulty score} of a training point $z_i$ is its loss with respect to the optimal hypothesis $\theta^*$, i.e., $\Psi_i = \ell(\theta^*, z_i)$.
\end{definition}

\begin{assumption}[Ideal Difficulty and Gradient Variance]\label{assump:low-difficulty-variance}
Consider SGD updates of the form~\eqref{eq:sgd_update} and define the stochastic gradient noise
\[
    \xi_t \;=\; g_t - \mathbb{E}[g_t \mid \theta_t],
    \qquad
    \sigma_t^2 \;=\; \mathbb{E}[\|\xi_t\|_2^2 \mid \theta_t].
\]
There exist constants $\sigma_{\mathrm{easy}}^2$ and $\sigma_{\mathrm{hard}}^2$ and a neighborhood $\mathcal{N}(\theta^*)$ such that
\[
    \sup_{\theta \in \mathcal{N}(\theta^*)} \, \mathbb{E}_{z \sim \mathcal{P}_{\mathrm{easy}}}\big[\|\nabla \ell(\theta,z) - \nabla F_{\mathrm{easy}}(\theta)\|_2^2\big]
    \;\le\; \sigma_{\mathrm{easy}}^2,
\]
\[
    \inf_{\theta \in \mathcal{N}(\theta^*)} \, \mathbb{E}_{z \sim \mathcal{P}_{\mathrm{hard}}}\big[\|\nabla \ell(\theta,z) - \nabla F_{\mathrm{hard}}(\theta)\|_2^2\big]
    \;\ge\; \sigma_{\mathrm{hard}}^2,
\]
with $\sigma_{\mathrm{hard}}^2 > \sigma_{\mathrm{easy}}^2$, where $F_{\mathrm{easy}}$ and $F_{\mathrm{hard}}$ denote the population risks restricted to $\mathcal{P}_{\mathrm{easy}}$ and $\mathcal{P}_{\mathrm{hard}}$, respectively.
\end{assumption}

\subsection{Uniform-sampling variance dynamics and score-based pacing}
We give the formal details underlying \S\ref{sec:theoretical_framework}.

\paragraph{Effective gradient variance under uniform sampling.}
To formalize the notion of ``effective'' gradient variance, consider uniform sampling from the full data distribution $\mathcal{P}$ and write it as a mixture of easy and hard components,
\[
    \mathcal{P} \;=\; (1-\rho)\,\mathcal{P}_{\mathrm{easy}} + \rho\,\mathcal{P}_{\mathrm{hard}},
    \qquad
    \rho \in (0,1).
\]
Let $\mu_{\mathrm{easy},t}=\nabla F_{\mathrm{easy}}(\theta_t)$ and $\mu_{\mathrm{hard},t}=\nabla F_{\mathrm{hard}}(\theta_t)$ denote the component-wise mean gradients, and let
\[
    v_{\mathrm{easy},t}
    =
    \mathbb{E}_{z \sim \mathcal{P}_{\mathrm{easy}}}
    \big[\|\nabla \ell(\theta_t,z)-\mu_{\mathrm{easy},t}\|_2^2\big],
    \qquad
    v_{\mathrm{hard},t}
    =
    \mathbb{E}_{z \sim \mathcal{P}_{\mathrm{hard}}}
    \big[\|\nabla \ell(\theta_t,z)-\mu_{\mathrm{hard},t}\|_2^2\big].
\]
By the law of total variance, the conditional gradient-noise variance under uniform sampling from this fixed mixture is
\[
    \sigma_t^2
    \;=\;
    (1-\rho)\,v_{\mathrm{easy},t}
    + \rho\,v_{\mathrm{hard},t}
    + \rho(1-\rho)\,\|\mu_{\mathrm{easy},t}-\mu_{\mathrm{hard},t}\|_2^2.
\]
Any time dependence in $\sigma_t^2$ comes from the optimization trajectory through the component variances and the separation between component mean gradients.

\begin{assumption}[Dynamic Increase of Effective Gradient Variance]\label{assump:dynamic-variance}
Under uniform sampling from $\mathcal{P}$, training proceeds so that easy-example gradients shrink earlier than hard-example gradients. Consequently, the fixed-mixture variance above shifts from a lower-variance regime to a higher-variance regime dominated by $v_{\mathrm{hard},t}$ and the between-component mean-gradient term. Holding gradient magnitude comparable, this scenario also raises the GNS.
\end{assumption}

\paragraph{Score-based pacing.}
To connect the stability analysis to concrete curricula, it is useful to model explicitly how data ordering is induced by a scalar score and a pacing function. Let $d$ be a scalar score on training samples, and let $D = \{z_{(i)}\}_{i=1}^N$ denote the training dataset sorted in non-decreasing order of $d(z_{(i)})$. Training progress is parameterized by $t \in [0,1]$, representing the fraction of total training steps.

\begin{definition}[Sorted-Stream Pacing Function]
Given a scalar score $d$, a sorted-stream pacing function is a nondecreasing mapping $q\colon [0,1] \to [0,1]$ that specifies the score-rank quantile visited at training progress $t$. For a local window width $h>0$, the samples encountered near time $t$ are idealized as
\[
    W_t
    =
    \left\{
        z_{(i)} \in D
        \;\middle|\;
        \left| \frac{i}{N} - q(t) \right| \le \frac{h}{2}
    \right\},
\]
with the endpoints clipped to $[0,1]$.
\end{definition}

The fully sorted one-pass curricula used in our experiments correspond to linear rank pacing, $q(t)=t$: early batches are drawn from low-score windows, and late batches are drawn from high-score windows. Concave schedules, such as $q(t)=\sqrt{t}$, reach higher-score windows earlier than linear pacing. Convex schedules, such as $q(t)=t^2$, remain longer in lower-score windows and delay exposure to high-score windows.

\subsection{Formal statement underlying the variance-control lens}
\begin{theorem}[Variance-control statement]
Given a scalar score $d$ and sorted-stream pacing function $q \colon [0,1] \to [0,1]$, let a curriculum present training samples in order of increasing $d$ so that the sampling distribution $\mathcal{P}_t$ at (rescaled) time $t$ is supported on a local score-rank window $W_t$ around $q(t)$. Assume that $F$ is $\mu$-strongly convex with $L$-Lipschitz gradient and that Assumptions~\ref{assump:low-difficulty-variance}--\ref{assump:dynamic-variance} hold. For SGD with constant step size $\eta \in (0,1/L]$:
\begin{enumerate}[leftmargin=*]
    \item There exists a stability threshold
    \[
        \sigma_{\mathrm{stab}}^2(R)
        \;=\;
        \frac{\mu}{\eta}\,R^2,
    \]
    parameterized by a target radius $R>0$, such that any sampling scheme whose effective gradient variance satisfies
    $\sup_t \sigma_t^2 \le \sigma_{\mathrm{stab}}^2(R)$ and initialization satisfies $\|\theta_0 - \theta^*\|_2^2 \le R^2$ guarantees
    \[
        \sup_t \mathbb{E}\big[\|\theta_t - \theta^*\|_2^2\big] \;\le\; R^2.
    \]
    \item Under uniform sampling from $\mathcal{P}$, let $\sigma_{\mathrm{unif,late}}^2$ denote the late-regime effective variance level implied by Assumption~\ref{assump:dynamic-variance}. If $\sigma_{\mathrm{unif,late}}^2 > \sigma_{\mathrm{stab}}^2(R)$, then the best bound obtainable from \Cref{lem:sgd_stability} has $R$ replaced by
    \[
        R_{\mathrm{unif}}^2
        \;=\;
        \frac{\eta}{\mu}\,\sigma_{\mathrm{unif,late}}^2
        \;>\; R^2.
    \]
    \item For a sorted-stream curriculum induced by $(d,q)$, early iterations visit low-score windows. If these windows satisfy the easy-region variance bound and the score-window variance stays below $\sigma_{\mathrm{stab}}^2(R)$ over an initial horizon $t \le T_{\mathrm{stab}}$, \Cref{lem:sgd_stability} implies
    \[
        \sup_{t \le T_{\mathrm{stab}}} \mathbb{E}\big[\|\theta_t - \theta^*\|_2^2\big] \;\le\; R^2.
    \]
\end{enumerate}
\end{theorem}

\subsection{SGD Stability Lemma}

\begin{lemma}[SGD Stability under Bounded Gradient Variance]\label{lem:sgd_stability}
Assume that $F$ is $\mu$-strongly convex and has $L$-Lipschitz gradient, and consider SGD updates of the form~\eqref{eq:sgd_update} with constant step size $\eta \in (0, 1/L]$. Suppose that for all $t$,
\[
    \mathbb{E}[g_t \mid \theta_t] = \nabla F(\theta_t)
    \quad\text{and}\quad
    \mathbb{E}[\|g_t - \nabla F(\theta_t)\|_2^2 \mid \theta_t] \le \sigma^2.
\]
Then the iterates satisfy
\begin{align}
    \mathbb{E}\big[\|\theta_t - \theta^*\|_2^2\big]
    \;\le\;
    & (1 - \mu \eta)^t \|\theta_0 - \theta^*\|_2^2 \nonumber \\
    & + \Big(1-(1-\mu\eta)^t\Big)\frac{\eta}{\mu}\,\sigma^2,
    \quad t \ge 0.
    \label{eq:sgd_bound}
\end{align}
where $\theta^*$ is the unique minimizer of $F$.
\end{lemma}

\subsection{Proofs for \Cref{lem:sgd_stability} and the variance-control statement}

\begin{proof}[Proof of \Cref{lem:sgd_stability}]
The proof follows standard SGD analysis. Taking conditional expectation given $\theta_t$,
\begin{align*}
    \mathbb{E}\big[\|\theta_{t+1} - \theta^*\|_2^2 \mid \theta_t\big]
    &= \mathbb{E}\big[\|\theta_t - \theta^* - \eta g_t\|_2^2 \mid \theta_t\big] \\
    &= \|\theta_t - \theta^* - \eta \nabla F(\theta_t)\|_2^2 \\
    &\quad + \eta^2\,\mathbb{E}\big[\|g_t - \nabla F(\theta_t)\|_2^2 \mid \theta_t\big],
\end{align*}
since the mixed term has zero conditional expectation under $\mathbb{E}[g_t \mid \theta_t]=\nabla F(\theta_t)$. For $\mu$-strongly convex $F$ with $L$-Lipschitz gradient and $\eta \in (0,1/L]$, a standard contraction bound gives
\[
    \|\theta_t - \theta^* - \eta \nabla F(\theta_t)\|_2^2
    \le (1-\mu\eta)\,\|\theta_t - \theta^*\|_2^2
    \qquad\text{for all $t$}
\]
(see, e.g.,~\citep{bottou2018optimization}). Combining with the variance bound yields
\[
    \mathbb{E}\big[\|\theta_{t+1} - \theta^*\|_2^2 \mid \theta_t\big]
    \le (1-\mu\eta)\,\|\theta_t - \theta^*\|_2^2 + \eta^2\sigma^2.
\]
Taking unconditional expectation and unrolling the recursion gives~\eqref{eq:sgd_bound}.
\end{proof}

\begin{proof}[Proof of the variance-control statement]
For any sampling scheme with $\sup_t \sigma_t^2 \le \bar{\sigma}^2$, \Cref{lem:sgd_stability} gives
\[
    \mathbb{E}\big[\|\theta_t - \theta^*\|_2^2\big]
    \le (1-\mu\eta)^t \|\theta_0 - \theta^*\|_2^2
       + \Big(1-(1-\mu\eta)^t\Big)\frac{\eta}{\mu}\,\bar{\sigma}^2,
\]
Choosing $\bar{\sigma}^2 = \sigma_{\mathrm{stab}}^2(R) = \tfrac{\mu}{\eta} R^2$ and assuming $\|\theta_0 - \theta^*\|_2^2 \le R^2$, the right-hand side is a convex combination of two quantities each at most $R^2$, yielding the claim in part (1).

For part (2), Assumption~\ref{assump:dynamic-variance} implies that under uniform sampling the effective variance reaches a late-regime level $\sigma_{\mathrm{unif,late}}^2$, so $\bar{\sigma}^2$ in the bound of \Cref{lem:sgd_stability} must be taken at least this large. The corresponding asymptotic radius is then $R_{\mathrm{unif}}^2 = (\eta/\mu)\sigma_{\mathrm{unif,late}}^2$, which exceeds $R^2$ when $\sigma_{\mathrm{unif,late}}^2 > \sigma_{\mathrm{stab}}^2(R)$.

For part (3), curricula induced by $(d,q)$ define a family of local score-window distributions $\mathcal{P}_t$ whose support is contained in $W_t$. For an ascending sorted stream, early windows lie in the low-score region of $D$. When these windows also satisfy the easy-region variance bound in Assumption~\ref{assump:low-difficulty-variance}, a pacing rule that keeps $\sup_{t \le T_{\mathrm{stab}}} \sigma_t^2$ below $\sigma_{\mathrm{stab}}^2(R)$ over an initial horizon permits applying \Cref{lem:sgd_stability} with $\bar{\sigma}^2 = \sigma_{\mathrm{stab}}^2(R)$, giving the stated bound.
\end{proof}

\begin{table}[t]
    \centering
    \scriptsize
    \begin{tabular}{llrrrrr}
    \toprule
    \textbf{Probe} & \textbf{Ordering} & \textbf{14M} & \textbf{31M} & \textbf{70M} & \textbf{160M} & \textbf{410M} \\
    \midrule
    \multirow{4}{*}{\emph{wh}-questions object-gap} 
     & VV & \cellcolor{rank1} \textbf{54.6} & \cellcolor{rank1} \textbf{74.2} & \cellcolor{rank1} \textbf{77.5} & \cellcolor{rank2} 76.2 & \cellcolor{rank4} 82.2 \\
     & AoA & \cellcolor{rank2} 53.1 & \cellcolor{rank3} 61.9 & \cellcolor{rank2} 70.1 & \cellcolor{rank1} \textbf{77.8} & \cellcolor{rank2} 83.7 \\
     & Frequency & \cellcolor{rank3} 42.9 & \cellcolor{rank2} 64.3 & \cellcolor{rank4} 61.6 & \cellcolor{rank3} 75.9 & \cellcolor{rank1} \textbf{85.4} \\
     & Random & \cellcolor{rank4} 42.4 & \cellcolor{rank4} 55.5 & \cellcolor{rank3} 63.9 & \cellcolor{rank4} 75.0 & \cellcolor{rank2} 83.7 \\
    \midrule
    \multirow{4}{*}{Causative} 
     & VV & \cellcolor{rank1} \textbf{66.5} & \cellcolor{rank1} \textbf{66.5} & \cellcolor{rank1} \textbf{73.9} & \cellcolor{rank1} \textbf{73.8} & \cellcolor{rank1} \textbf{78.3} \\
     & AoA & \cellcolor{rank3} 61.6 & \cellcolor{rank3} 64.0 & \cellcolor{rank2} 68.3 & \cellcolor{rank2} 72.8 & \cellcolor{rank3} 76.3 \\
     & Frequency & \cellcolor{rank2} 61.8 & \cellcolor{rank4} 62.2 & \cellcolor{rank4} 65.8 & \cellcolor{rank3} 72.3 & \cellcolor{rank2} 77.7 \\
     & Random & \cellcolor{rank4} 58.5 & \cellcolor{rank2} 64.4 & \cellcolor{rank3} 67.5 & \cellcolor{rank4} 71.6 & \cellcolor{rank4} 74.8 \\
    \midrule
    \multirow{4}{*}{Superlative quantifiers} 
     & VV & \cellcolor{rank3} 31.4 & \cellcolor{rank4} 47.4 & \cellcolor{rank2} 71.5 & \cellcolor{rank4} 77.6 & \cellcolor{rank2} 86.6 \\
     & AoA & \cellcolor{rank4} 26.3 & \cellcolor{rank1} \textbf{75.1} & \cellcolor{rank3} 65.6 & \cellcolor{rank2} 87.6 & \cellcolor{rank4} 81.7 \\
     & Frequency & \cellcolor{rank2} 35.5 & \cellcolor{rank3} 62.6 & \cellcolor{rank4} 43.8 & \cellcolor{rank3} 80.5 & \cellcolor{rank1} \textbf{89.1} \\
     & Random & \cellcolor{rank1} \textbf{61.9} & \cellcolor{rank2} 63.1 & \cellcolor{rank1} \textbf{75.5} & \cellcolor{rank1} \textbf{95.3} & \cellcolor{rank3} 85.6 \\
    \midrule
    \multirow{4}{*}{Only-NPI scope} 
     & VV & \cellcolor{rank3} 58.7 & \cellcolor{rank2} 67.6 & \cellcolor{rank3} 48.8 & \cellcolor{rank1} \textbf{76.6} & \cellcolor{rank3} 70.9 \\
     & AoA & \cellcolor{rank4} 50.4 & \cellcolor{rank4} 56.5 & \cellcolor{rank1} \textbf{67.2} & \cellcolor{rank3} 61.7 & \cellcolor{rank2} 76.0 \\
     & Frequency & \cellcolor{rank2} 58.9 & \cellcolor{rank1} \textbf{70.0} & \cellcolor{rank4} 26.6 & \cellcolor{rank2} 71.7 & \cellcolor{rank4} 64.0 \\
     & Random & \cellcolor{rank1} \textbf{61.3} & \cellcolor{rank3} 58.0 & \cellcolor{rank2} 60.8 & \cellcolor{rank4} 54.4 & \cellcolor{rank1} \textbf{78.7} \\
    \midrule
    \multirow{4}{*}{Ellipsis N-bar} 
     & VV & \cellcolor{rank1} \textbf{61.4} & \cellcolor{rank1} \textbf{69.5} & \cellcolor{rank1} \textbf{72.8} & \cellcolor{rank1} \textbf{83.7} & \cellcolor{rank3} 85.3 \\
     & AoA & \cellcolor{rank4} 54.1 & \cellcolor{rank2} 62.9 & \cellcolor{rank3} 70.0 & \cellcolor{rank3} 80.7 & \cellcolor{rank2} 85.9 \\
     & Frequency & \cellcolor{rank2} 57.1 & \cellcolor{rank4} 58.9 & \cellcolor{rank4} 65.9 & \cellcolor{rank4} 77.6 & \cellcolor{rank4} 81.7 \\
     & Random & \cellcolor{rank3} 56.8 & \cellcolor{rank3} 61.2 & \cellcolor{rank2} 70.1 & \cellcolor{rank2} 82.7 & \cellcolor{rank1} \textbf{87.0} \\
    \bottomrule
    \end{tabular}
    \caption{Final BLiMP accuracy (\%) at 300B tokens for selected probes. Best result per probe and model size in bold.}
    \label{tab:blimp_summary}
\end{table}

\begin{figure}[t]
    \centering
    \begin{subfigure}{\linewidth}
        \centering
        \includegraphics[width=\linewidth]{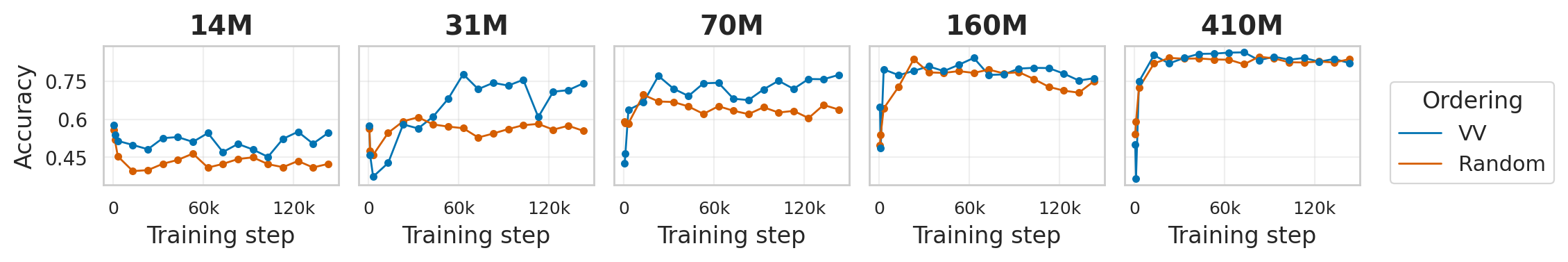}
        \caption{VV vs Random on \emph{wh}-questions object-gap. VV (blue) outperforms Random (orange) most clearly at smaller scales.}
        \label{fig:blimp_wh_pairwise}
    \end{subfigure}
    \begin{subfigure}{\linewidth}
        \centering
        \includegraphics[width=\linewidth]{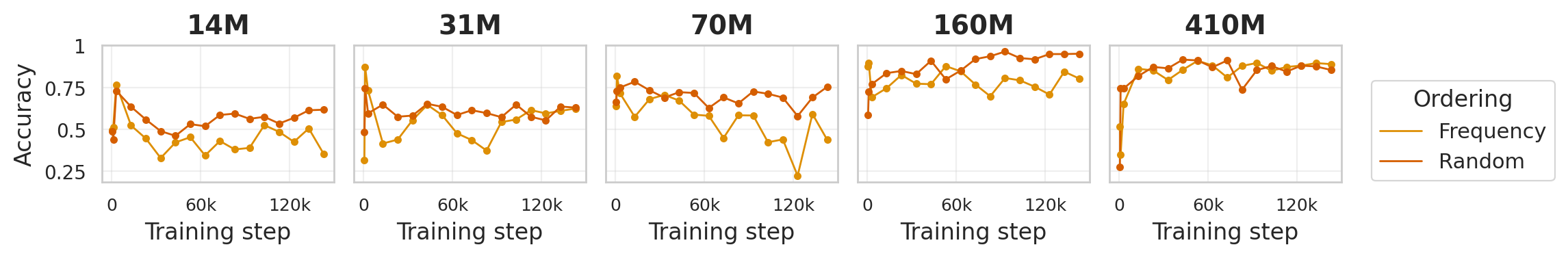}
        \caption{Frequency vs Random on superlative quantifiers. Random (orange) often outperforms Frequency (blue), with high variance across model sizes.}
        \label{fig:blimp_superlative_pairwise}
    \end{subfigure}
    \caption{Pairwise ordering comparisons on BLiMP probes with recurring differences. \subref{fig:blimp_wh_pairwise} shows VV versus Random on \emph{wh}-questions object-gap; \subref{fig:blimp_superlative_pairwise} shows Frequency versus Random on superlative quantifiers.}
    \label{fig:blimp_pairwise}
\end{figure}

\section{Reverse-Order Control}
\label{sec:appendix_reverse_control}

\begin{wrapfigure}[14]{r}{0.58\textwidth}
    \vspace{-25pt}
    \centering
    \begin{subfigure}{0.49\linewidth}
        \centering
        \includegraphics[width=\linewidth]{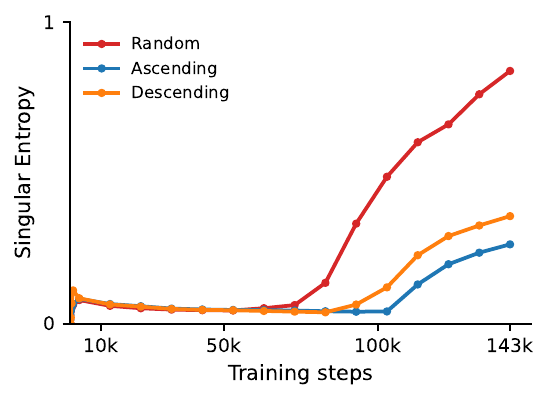}
        \caption{Singular entropy.}
        \label{fig:reverse_entropy}
    \end{subfigure}
    \hfill
    \begin{subfigure}{0.49\linewidth}
        \centering
        \includegraphics[width=\linewidth]{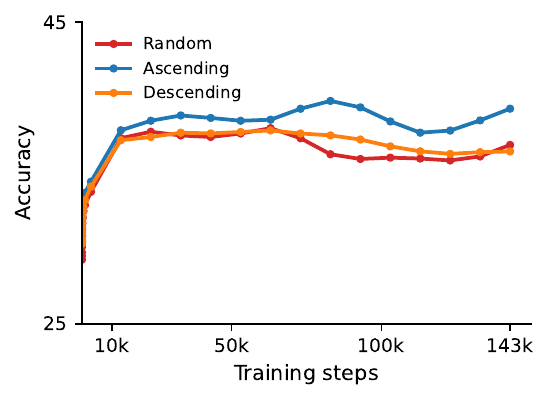}
        \caption{Evaluation accuracy.}
        \label{fig:reverse_accuracy}
    \end{subfigure}
    \caption{Reverse-order control for the VV score using the 31M model trained for 300B tokens. \subref{fig:reverse_entropy} shows singular entropy; \subref{fig:reverse_accuracy} shows evaluation accuracy. Descending order reduces accuracy relative to ascending VV, while its late singular-entropy trajectory remains below Random.}
    \label{fig:reverse_control}
\end{wrapfigure}

Figure~\ref{fig:reverse_control} reports a targeted reverse-order control using the 31M model trained for 300B tokens with the VV ordering. The ascending condition is the ascending VV-score curriculum used in the main experiments, while the descending condition presents the same VV-scored samples in reverse order. Following the evaluation in \citet{biderman2023pythia}, the plotted accuracy is the mean over the eight benchmarks: LAMBADA, PiQA, WinoGrande, WSC, ARC Easy, ARC Challenge, SciQ, and LogiQA. Reversing the order reduces accuracy relative to ascending VV through most of training and removes the late-training accuracy advantage of the ascending run. For singular entropy, the descending run rises earlier and higher than ascending VV, but remains below Random at the end of training. This control shows that VV ordering direction matters, while leaving open the broader domain-mixture and proxy-difficulty caveats discussed in the main text.

\section{BLiMP Trajectories by Curriculum Ordering}
\label{sec:appendix_shuffle_blimp}

This appendix reports the BLiMP trajectories behind the ordering effects discussed in \S\ref{sec:results}. We compare each non-random ordering to Random across model sizes and focus on probes with differences that meet our selection thresholds. Table~\ref{tab:blimp_summary} summarizes final accuracy for selected probes; Figure~\ref{fig:blimp_pairwise} shows pairwise training trajectories for the key comparisons.

\section{Computational Costs} \label{sec:compute_costs}

\begin{wraptable}[9]{r}{0.4\textwidth}
    \vspace{-20pt}
    \centering
    \begin{tabular}{lrr}
    \toprule
    Model & A100 & 2080 Ti \\
    \midrule
    14M & 320 & 860 \\
    31M & 350 & 1170 \\
    70M & 530 & --- \\
    160M & 1140 & --- \\
    410M & 2730 & --- \\
    1B & 5700 & --- \\
    \bottomrule
    \end{tabular}
    \caption{Computational costs (in GPU-hours) for pretraining.}
    \label{tab:compute_costs}
\end{wraptable}

Experiments on 14M and 31M models were conducted primarily using 64 2080 Ti GPUs, while experiments on 70M, 160M, and 410M models were conducted using 16 80GB NVIDIA A100 GPUs. The computational cost in GPU-hours is provided in Table~\ref{tab:compute_costs}. Our measured GPU-hours are comparable to those in the original Pythia paper: for 70M, 160M, and 410M, \citet{biderman2023pythia} report 510, 1030, and 2540 GPU-hours, respectively.

\section{Curriculum Design and Scoring Details} \label{sec:curriculum_details}

\subsection{Scoring Functions}
Since our chosen curriculum metrics are word-level, each 2048-token sample must first be detokenized to produce a word sequence before scoring. For a given detokenized text sequence $S$ containing $N$ words $\{w_1, \dots, w_N\}$, we define the curriculum scores as follows:
\begin{itemize}[leftmargin=*]
    \item \textbf{Age-of-Acquisition (AoA)}: This score is based on the Kuperman et al. age of acquisition norms~\citep{kuperman2012age}, representing the age at which a word is typically learned. The sequence score is the average AoA of its constituent words:
    \[ \text{Score}_{\text{AoA}}(S) = \frac{1}{N} \sum_{i=1}^{N} \text{AoA}(w_i) \]
    \item \textbf{Word Frequency (Freq)}: This score is based on the SUBTLEX-US word frequency database~\citep{brysbaert2009subtlex}, using the Zipf scale~\citep{van2014subtlex}. The sequence score is the average Zipf value of its words:
    \[ \text{Score}_{\text{Freq}}(S) = \frac{1}{N} \sum_{i=1}^{N} \text{Zipf}(w_i) \]
    \item \textbf{Verb Variation (VV)}: This score measures the linguistic diversity of verbs within a sample. It is computed as the number of unique verbs divided by the square root of the total number of verbs:
    \[ \text{Score}_{\text{VV}}(S) = \begin{cases} \dfrac{\text{Unique Verbs}}{\sqrt{\text{Total Verbs}}} &  \text{if Total Verbs} \ne 0 \\ 0 & \text{if Total Verbs} = 0 \end{cases} \]
    The square root adjustment reduces dependence on the total number of detected verbs.
\end{itemize}
Normalizing by the number of words $N$ mitigates length bias. For our curricula, we sort samples in ascending AoA, ascending Frequency (Zipf), and ascending VV; higher-score samples tend to have higher empirical sample loss under trained checkpoints (Table~\ref{tab:correlation_stats}). The linguistic features were computed using the LFTK toolkit~\citep{lee-lee-2023-lftk}, which relies on spaCy's English pipeline (en\_core\_web\_sm) and English lexical resources (Kuperman AoA; SUBTLEX Zipf). Because The Pile contains code, numeric tables, and non-English text, some tokens fall outside these lexical resources and receive default values of 0 in the toolkit, which can affect the extreme low-score tails. We find the resulting scores remain strongly correlated with empirical sample loss across model sizes (Table~\ref{tab:correlation_stats}).

\subsection{Sample Splitting and Concatenation}
\label{sec:sample_split}

\begin{wrapfigure}[10]{r}{0.6\textwidth}
    \vspace{-15pt}
    \centering
    \includegraphics[width=\linewidth]{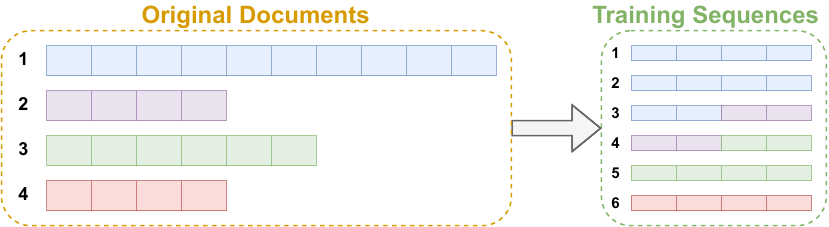}
    \caption{Illustration of the sequence packing process used by Pythia. Smaller documents are concatenated, while larger documents are split across multiple 2048-token samples.}
    \label{fig:sample_split}
\end{wrapfigure}

We follow the data processing methodology of Pythia~\citep{biderman2023pythia}, which employs ``sequence packing''~\citep{raffel2020exploring} to create a dataset of uniform-length training samples. Documents from The Pile are processed into fixed-size, 2048-token sequences. As illustrated in Figure~\ref{fig:sample_split}, this process concatenates smaller documents and splits larger ones across multiple sequences to maximize computational efficiency. Because every curriculum reuses this same packed sample set, the intervention changes only the order of samples, not the underlying token sequences.

\subsection{Correlation with Sample Loss}
\label{sec:correlation_analysis}

\begin{wraptable}[12]{r}{0.45\textwidth}
    \vspace{-10pt}
    \centering
    \begin{tabular}{lccc}
    \toprule
    Model & Freq. & AoA & VV \\
    \midrule
    14M & 0.717 & 0.611 & 0.780 \\
    31M & 0.723 & 0.611 & 0.773 \\
    70M & 0.734 & 0.618 & 0.768 \\
    160M & 0.734 & 0.615 & 0.755 \\
    410M & 0.745 & 0.627 & 0.746 \\
    \midrule
    \textbf{Average} & \textbf{0.730} & \textbf{0.616} & \textbf{0.764} \\
    \bottomrule
    \end{tabular}
    \caption{Pearson correlation of linguistic indices with sample loss.}
    \label{tab:correlation_stats}
\end{wraptable}

We computed the Pearson correlation between our sample scores (Average AoA, Average Word Frequency, and VV) and the average sample loss. To compute this, we sampled 100,000 sequences from the dataset, evaluated their losses on all checkpoints of all models from 14M to 410M, and averaged the results. As summarized in Table~\ref{tab:correlation_stats}, correlations are consistently positive for all three metrics across model sizes, showing that the linguistic scores are aligned with empirical sample loss in our setting. We use this empirical sample loss as a proxy for the ideal difficulty $\Psi_i=\ell(\theta^*,z_i)$, but it is not identical to $\Psi_i$: it is measured under finite trained checkpoints, and it can reflect model capacity, training stage, corpus domain, and lexical-resource coverage as well as intrinsic sample difficulty.

\subsection{Domain--Difficulty Confounding}
\label{sec:domain_difficulty_confound}

\begin{figure*}[t]
    \centering
    \begin{subfigure}{0.49\linewidth}
        \centering
        \includegraphics[width=\linewidth]{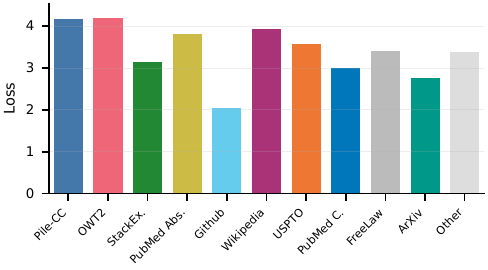}
        \caption{Average empirical sample loss by Pile source domain.}
        \label{fig:domain_loss}
    \end{subfigure}
    \begin{subfigure}{0.49\linewidth}
        \centering
        \includegraphics[width=\linewidth]{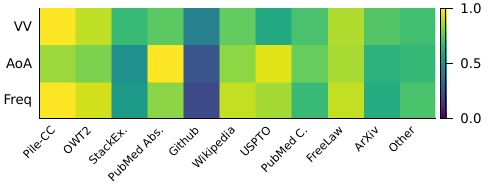}
        \caption{Domain-level average curriculum scores, normalized within each score.}
        \label{fig:domain_score_heatmap}
    \end{subfigure}
    \caption{Domain-level evidence that linguistic-score sorting is entangled with source-domain exposure. Domains differ substantially in empirical sample loss: GitHub and ArXiv are lower-loss on average, while Pile-CC, OpenWebText2, Wikipedia, and PubMed Abstracts are higher-loss. The same domains also occupy different regions of the AoA, Frequency, and VV score ranges. Sorting by these scores therefore changes both the empirical difficulty profile and the domain mixture seen during training.}
    \label{fig:domain_difficulty_confound}
\end{figure*}

Figure~\ref{fig:domain_difficulty_confound} shows a limitation of interpreting the linguistic curricula as pure difficulty interventions. The sample-loss correlations in Table~\ref{tab:correlation_stats} show that the scores are aligned with empirical loss, but the domain-level analysis shows that part of this alignment is mediated by source domain. For example, GitHub has low average loss and low normalized scores across all three curricula, while Pile-CC and OpenWebText2 have high average loss and high VV/Frequency scores. AoA assigns especially high scores to PubMed Abstracts and USPTO, reflecting domain-specific vocabulary and lexical-resource coverage rather than a domain-invariant notion of difficulty.

Consequently, the GNS and singular-entropy differences in the main experiments should be read as effects of linguistic-score orderings over The Pile, where difficulty, register, and source domain move together. This does not invalidate the ordering effect: all runs use the same fixed samples, and the intervention is still the order in which those samples are presented. It does mean that the present experiments cannot determine whether lower GNS and weaker late-stage spectral saturation come from difficulty pacing itself, from changes in domain exposure, or from their interaction. A clean separation would require domain-stratified or domain-matched curricula, for example sorting within each Pile source or matching curriculum windows by source-domain composition.

\subsection{Dataset Excerpts Across Curriculum Quantiles}
\label{sec:curated_excerpts}

We provide excerpts from the quantile inspection set used in \S\ref{sec:linguistic_characterization}. For word frequency, the low-frequency prefix often includes non-standard content such as source code, structured data, URLs, and non-English text, alongside technical writing with dense domain-specific terminology; mid-to-high frequency quantiles include narrative and conversational text as well as academic prose; and the extreme high-frequency tail can become dominated by repetitive strings. Consistent with Table~\ref{tab:correlation_stats}, lower-frequency quantiles tend to have lower sample loss, while higher-frequency quantiles tend to have higher loss. AoA produces a related ordering: low-AoA regions include conversational or instructional English and non-linguistic content (e.g., code, numeric tables) when lexical coverage is limited; mid-range AoA samples include formal prose and documentation; higher-AoA samples often feature specialized academic or technical vocabulary. VV orders samples by detected verb-type diversity: low-VV quantiles include samples with few detected verbs (markup, lists, code); mid-range VV samples include formal expository prose; and the highest-VV tail frequently includes verb lexicons that inflate type counts. Table~\ref{tab:curriculum_quantile_excerpts} gives representative examples.

\begin{table*}[t]
    \centering
    \small
    \setlength{\extrarowheight}{1pt}
    \renewcommand{\arraystretch}{1.05}
    \begin{tabularx}{\textwidth}{@{} >{\centering\arraybackslash}p{3.5em} @{\hspace{6pt}} P{11.5em} @{\hspace{6pt}} X @{}}
    \toprule
    \multicolumn{3}{@{}l}{\textbf{Frequency}} \\
    \midrule
    \rowcolor{qLow}
    \textbf{25\%} & Expository prose &
    \detokenize{Some adults seem not to fear heights}\newline
    \detokenize{Training instils confidence that you will not fall}\newline
    \detokenize{Tests of crawling infants have shown them to avoid an apparent (but glass-covered) precipice.}
    \\
    \rowcolor{qMid}
    \textbf{50\%} & Scientific prose &
    \detokenize{Ladybirds (Coleoptera: Coccinellidae) are a diverse group of beetles that are usually brightly coloured.}
    \\
    \rowcolor{qHigh}
    \textbf{75\%} & Narrative reference &
    \detokenize{The film series will continue with an untitled third Kingsman film and a spin-off film Statesman in development.}\newline
    \detokenize{The franchise will also expand to television with an eight-hour limited series in development.}
    \\
    \rowcolor{qTail}
    \textbf{100\%} & Repetitive / degenerate tail &
    \detokenize{A A A A A A A A A A A A A A A A A A A A A A A A A A ...}
    \\
    \midrule
    \multicolumn{3}{@{}l}{\textbf{Age-of-Acquisition}} \\
    \midrule
    \rowcolor{qLow}
    \textbf{25\%} & Dialogue / transcript-like &
    \detokenize{"Opening." "Welcome to the Montecito."}\newline
    \detokenize{[Crowd Chattering, Cheering]}\newline
    \detokenize{"All right, everybody, here we go."}
    \\
    \rowcolor{qMid}
    \textbf{75\%} & Forum-style question &
    \detokenize{Is it OK to purchase cat food?}\newline
    \detokenize{Forgive a (possibly) silly question - but I have a cat as a pet, and of course I purchase cat food for him and feed it to him.}
    \\
    \rowcolor{qHigh}
    \textbf{100\%} & Keyword-stuffed tail &
    \detokenize{life insurance massachusetts long term care insurance illinois term life insurance quote insurance life life term whole health insurance ...}
    \\
    \midrule
    \multicolumn{3}{@{}l}{\textbf{Verb Variation}} \\
    \midrule
    \rowcolor{qLow}
    \textbf{0\%} & Markup / lists &
    \detokenize{<li class="nav-group-task">}\newline
    \detokenize{  <a class="nav-group-task-link" href="..."> MBVisaRecognizerResult</a>}\newline
    \detokenize{</li>}
    \\
    \rowcolor{qMid}
    \textbf{50\%} & Expository prose &
    \detokenize{To recognise always that the test of police efficiency is the absence of crime and disorder,}\newline
    \detokenize{and not the visible evidence of police action in dealing with them.}
    \\
    \rowcolor{qHigh}
    \textbf{100\%} & Verb lexicon / translation pairs &
    \detokenize{bautizar _to baptize, name_}\newline
    \detokenize{beber _to drink_}\newline
    \detokenize{bendecir _to bless_}\newline
    \detokenize{beneficiar _to benefit, sell at a discount_}
    \\
    \bottomrule
    \end{tabularx}
    \caption{Representative excerpts from curriculum-score quantiles. Frequency and AoA tails can include repetitive or keyword-stuffed text, while the high-VV tail can include verb lexicons that inflate detected verb-type diversity. URLs may be lightly redacted for readability.}
    \label{tab:curriculum_quantile_excerpts}
\end{table*}

\subsection{The Zipf Scale for Word Frequency}
The Zipf scale~\citep{van2014subtlex} is the logarithmic word-frequency scale used for the Frequency curriculum in \S\ref{sec:method}. It is calculated as the base-10 logarithm of frequency per billion words, equivalently $\log_{10}(\text{fpmw}) + 3$ for frequency per million words. This yields a compact scale, typically 1--7, that is easier to interpret than raw counts.

\section{Details of Probed Metrics}
\label{sec:probed_metrics_details}

The evaluation suite includes science question answering, long-range context comprehension, logical reasoning, and commonsense understanding. Specifically, the AI2 Reasoning Challenge (ARC)~\citep{clark2018think} assesses science question answering using both the Easy and Challenge subsets. LAMBADA~\citep{paperno2016lambada} targets long-range context comprehension by requiring prediction of the final word in narrative passages. LogiQA~\citep{liu2020logiqa} evaluates logical reasoning in reading comprehension with multiple-choice questions adapted from LSAT exams. PiQA~\citep{bisk2020piqa} measures commonsense physical reasoning via multiple-choice format. SciQ~\citep{welbl2017crowdsourcing} focuses on scientific question answering from educational materials. The Winograd Schema Challenge (WSC)~\citep{levesque2012winograd} tests commonsense pronoun resolution in minimally differing sentence pairs, and WinoGrande~\citep{sakaguchi2021winogrande} extends this setting with an emphasis on robustness to dataset biases. Accuracy is the primary reported metric, with language-model log-likelihood used where benchmark scoring requires it.

\section{HMM Weight Metrics}
\label{sec:hmm_metrics}

\begin{table}[t]
    \centering
    \setlength{\tabcolsep}{3pt}
    \renewcommand{\arraystretch}{0.92}
    \begin{tabular}{l l}
    \toprule
    Name & Description \\
    \midrule
    \multicolumn{2}{@{}l}{\textbf{Weight dispersion metrics}} \\
    $L_{1}$ & Average $L_{1}$-norm over all weight matrices. \\
    $L_{2}$ & Average $L_{2}$-norm over all weight matrices. \\
    $L_{1}/L_{2}$ & Average ratio of $L_{1}$ to $L_{2}$ norm. \\
    $\mu(w)$ & Sample mean of all scalar weights. \\
    $\mathrm{median}(w)$ & Median over all scalar weights. \\
    $\widehat{\mathrm{Var}}(w)$ & Sample variance of all scalar weights. \\
    $\mu(b)$ & Sample mean of all scalar biases. \\
    $\mathrm{median}(b)$ & Median over all scalar biases. \\
    $\widehat{\mathrm{Var}}(b)$ & Sample variance of all scalar biases. \\
    \midrule
    \multicolumn{2}{@{}l}{\textbf{Layer function metrics}} \\
    trace & Average trace over all weight matrices. \\
    $\lambda_{\max}$ & Average spectral norm over all weight matrices. \\
    trace/$\lambda_{\max}$ & Average trace-to-spectral-norm ratio. \\
    $\mu(\lambda)$ & Average singular value over all matrices. \\
    $\widehat{\mathrm{Var}}(\lambda)$ & Sample variance of singular values. \\
    \bottomrule
    \end{tabular}
    \caption{The 14 checkpoint-level metrics used as HMM observations.}
    \label{tab:hmm_metrics}
\end{table}

\begin{wrapfigure}[19]{r}{0.45\textwidth}
    \vspace{-10pt}
    \centering
    \includegraphics[width=\linewidth]{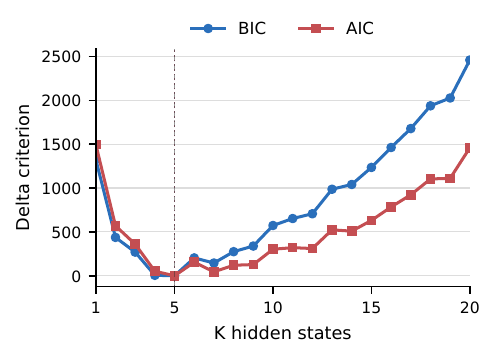}
    \caption{Sensitivity of HMM model selection to the number of hidden states. The plot shows the change in BIC and AIC relative to the best value across \(K\in\{1,\ldots,20\}\). Both criteria select \(K=5\), supporting the five-state training map used in Figure~\ref{fig:hmm_overview}.}
    \label{fig:hmm_k_sensitivity}
\end{wrapfigure}
Following the methodology of~\citet{hu2023latent}, we compute a set of 14 metrics from the network's weights at each checkpoint to serve as the observation sequence for the HMM.\footnote{The implementation for computing these metrics is adapted from \url{https://github.com/michahu/modeling-training/}.} These metrics capture properties of the weight distribution and the function computed by each layer. The full list is provided in Table~\ref{tab:hmm_metrics}. The first group tracks weight dispersion through norms, moments, and sparsity-like ratios; the second group tracks layer-function summaries through trace and singular-value statistics. Before fitting the HMM, each metric trajectory is standardized so that large-scale quantities do not dominate the observation space. Using these checkpoint-level summaries avoids fitting directly in the full parameter space while preserving the coarse information needed to identify shared training phases across orderings. These metrics summarize parameter scale and spectral structure for the HMM.

Figure~\ref{fig:hmm_k_sensitivity} reports the state-count sensitivity sweep used to choose the five-state HMM in the main phase analysis. Both BIC and AIC select \(K=5\), matching the phase resolution used in Figure~\ref{fig:hmm_overview}. We use this criterion only to set the coarse phase resolution; the downstream comparisons then hold the fitted state space fixed across orderings. This keeps the HMM analysis focused on differences in phase occupancy and within-phase exposure rather than on changing the number of latent states per curriculum.


\end{document}